\documentclass[11pt,a4paper]{article}
\usepackage[hyperref]{naaclhlt2019}
\usepackage{times}
\usepackage{latexsym}

\usepackage{url}
\usepackage{graphicx}  %Required
\usepackage{subfigure}
\usepackage{amsmath}
\usepackage{multirow}
\usepackage{subfigure}
\usepackage{arydshln}
\usepackage{color}
\usepackage{amsmath,amssymb,bm}
\usepackage{algorithm}
\usepackage{algorithmic}
\usepackage[american]{babel}
\usepackage[c]{esvect}
\usepackage{footnote}
\usepackage{enumerate}
\usepackage{CJKutf8}
\usepackage{footnote}

\aclfinalcopy % Uncomment this line for the final submission
\title{Syntax-Enhanced Neural Machine Translation with Syntax-Aware Word Representations}

\author{Meishan Zhang$^1$ ~~\textnormal{and}~~ Zhenghua Li$^2$ ~~\textnormal{and}~~ Guohong Fu$^3$\thanks{~~Corresponding author.}~~\textnormal{and}~~ Min Zhang$^2$ \\
1. School of New Media and Communication, Tianjin University, China \\
2. School of Computer Science and Technology, Soochow University, China \\
3. Institute of Artificial Intelligence, Soochow University, China \\
{\tt mason.zms@gmail.com,} \\
{\tt \{zhli13, minzhang\}@suda.edu.cn,} \\
{\tt ghfu@hotmail.com }
}

\date{}

\hypersetup{draft}
\begin{document}

\begin{CJK}{UTF8}{gbsn}

\maketitle
\begin{abstract}
Syntax has been demonstrated highly effective in neural machine translation (NMT).
Previous NMT models integrate syntax by representing 1-best tree outputs from a well-trained parsing system,
e.g., the representative Tree-RNN and Tree-Linearization methods, which may suffer from error propagation.
In this work, we propose a novel method to integrate source-side syntax implicitly for NMT.
The basic idea is to use the intermediate hidden representations of a well-trained end-to-end dependency parser,
which are referred to as syntax-aware word representations (SAWRs).
Then, we simply concatenate such SAWRs with ordinary word embeddings to enhance basic NMT models.
The method can be straightforwardly integrated into the widely-used sequence-to-sequence (Seq2Seq) NMT models.
We start with a representative RNN-based Seq2Seq baseline system,
and test the effectiveness of our proposed method on two benchmark datasets of the Chinese-English and English-Vietnamese translation tasks, respectively.
Experimental results show that the proposed approach is able to bring significant BLEU score improvements on the two datasets compared with the baseline,
1.74 points for Chinese-English translation and 0.80 point for English-Vietnamese translation, respectively.
In addition, the approach also outperforms the explicit Tree-RNN and Tree-Linearization methods.
\end{abstract}

\section{Introduction}

In the past few years, neural machine translation (NMT) has drawn increasing interests
due to its simplicity and promising performance \cite{bahdanau2014neural,jean2015montreal,stanfordmt2015,luong-pham-manning:2015:EMNLP,shen-EtAl:2016:P16-1,vaswani2017attention}.
%Especially,
The widely used
%Representative models exploit
sequence-to-sequence (Seq2Seq) framework combined with attention mechanism achieves
significant improvement over the traditional statistical machine translation (SMT) models
on a variety of language pairs, such as Chinese-English \cite{shi-padhi-knight:2016:EMNLP2016,mi-EtAl:2016:EMNLP2016,vaswani2017attention,yongcheng:2018:ACLMain}.
Under an encoder-decoder architecture,
the Seq2Seq framework first encodes the source sentence  into a sequence of hidden vectors,
and then incrementally predicts the target sentence \cite{cho-EtAl:2014:SSST-8}.

\begin{figure}[tb]
	\begin{center}
		\subfigure{ \label{fig:dependency}
			\centering{\includegraphics[scale=0.85]{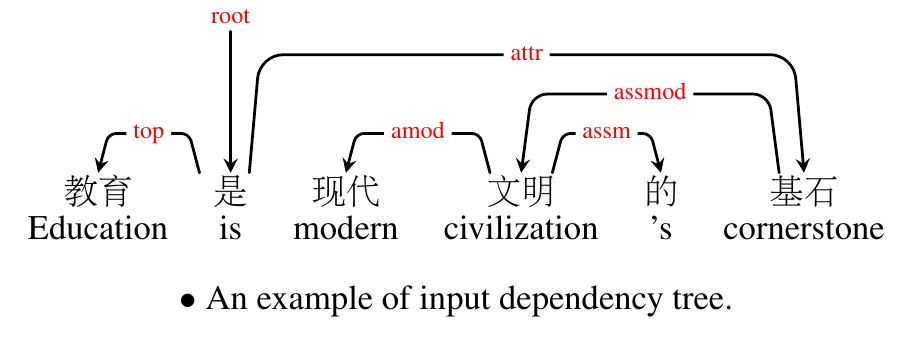}}			
		}
%\vspace{-0.5cm}		
		%\subfigure[Baseline source input without encoding dependency syntax.]{\label{fig:bi:baseline}
		%	\centering{\includegraphics[scale=1.0]{figures/figure1-baseline.pdf}}
		%}
		\subfigure{\label{fig:parser}
			\centering{\includegraphics[scale=1.05]{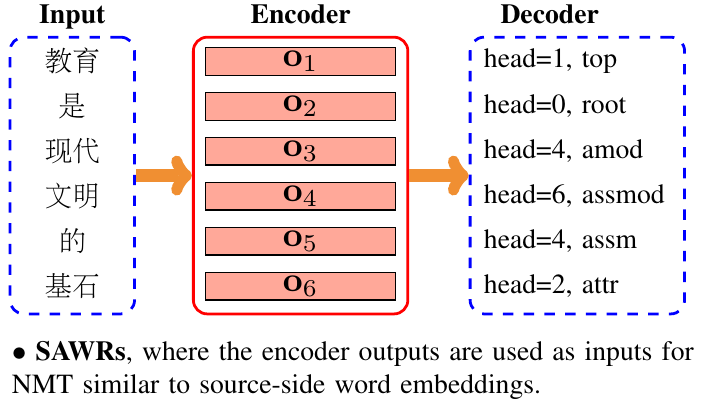}}
		}
		\caption{An example to illustrate our method of encoding source dependency syntax, where the English translation is ``Education is the cornerstone of modern civilization'' for the source Chinese input.}\label{fig:overall:method}
	\end{center}
%\vspace{-0.2cm}
\end{figure}

Recently, inspired by the success of syntax-based SMT \cite{williams2016syntax},
%Syntax information has been demonstrated highly effective in syntax-based SMT models \cite{williams2016syntax}.
%In the community of NMT,
researchers propose a range of interesting approaches for exploiting syntax information in NMT models,
%syntax information has also been explored intensively,
 as syntactic trees could offer long-distance relations in sentences
\cite{shi-padhi-knight:2016:EMNLP2016,wu2017improved,li-EtAl:2017:Long,bastings-EtAl:2017:EMNLP2017,hashimoto-tsuruoka:2017:EMNLP2017}.

As a straightforward method, tree-structured recurrent neural network (\textbf{Tree-RNN})
can elegantly model the source-side syntax and globally encode the whole trees. %is to %represent syntactic trees directly
%by recursive neural networks (RNN).
%based on composing methods such as gated recurrent unit (GRU) or long short term memory (LSTM),
%encoding the whole trees globally
\newcite{eriguchi-hashimoto-tsuruoka:2016:P16-1}, \newcite{chen-EtAl:2017:Long6} and \newcite{yang-EtAl:2017:EMNLP20172}
show that Tree-RNN can effectively integrate syntax-oriented trees into Seq2Seq NMT models.   %\cite{eriguchi-hashimoto-tsuruoka:2016:P16-1,chen-EtAl:2017:Long6,yang-EtAl:2017:EMNLP20172}.

Regardless of the effectiveness of Tree-RNN,
we find that it suffers from a severe low-efficiency problem
because of the heterogeneity of different syntax trees,
which leads to increasing difficulties for batch computation
compared with sequential inputs.
Even with deliberate batching method of \newcite{neubig17nips},
our preliminary experiments show that Tree-RNN
with gated recurrent unit (GRU) can lead to nearly four times slower performance
when it is integrated into a classical Seq2Seq system. %proposed by \newcite{bahdanau2014neural}.

To solve the problem, \textbf{Tree-Linearization} is a good alternative for syntax encoding.
The main idea is to linearize syntax trees into sequential symbols,
and then exploit the resulting sequences as inputs for NMT.
\newcite{li-EtAl:2017:Long} propose a depth-first method to traverse a constituent tree,
converting it into a sequence of symbols mixed with sentential words and syntax labels.
Similarly, \newcite{wu2017improved} combine several strategies of tree traversing for dependency syntax integration.

%All the above methods attempt to encode syntax by modeling discrete 1-best tree outputs of syntax parsers,
%which may cause an error propagation problem.
%Input errors produced by syntactic parsers may further propagate into automatic NMT results,
%leading to poor translation performance,
%since the input trees are directly associated with NMT outputs.
%The problem could make syntax less useful for NMT.
%One possible solution is to avoid the use of final outputs of syntax parsers as inputs for NMT.

%All the above methods attempt to encode syntax by modeling discrete 1-best tree outputs of syntax parsers,
%which may cause an error propagation problem.
In this work, we present an implicit syntax encoding method for NMT,
enhancing NMT models by syntax-aware word representations (\textbf{SAWRs}).
Figure \ref{fig:overall:method} illustrates the basic idea,
where trees are modeled indirectly by sequential vectors extracted from an encoder-decoder dependency parser.
On the one hand, the method avoids the structural heterogeneity and thus can be integrated efficiently,
and on the other hand,
it does not require discrete 1-best tree outputs,
alleviating the error propagation problem induced from syntax parsers.
Concretely,
the vector outputs are extracted from the encoding outputs of the encoder-decoder dependency parser.
As shown in Figure \ref{fig:overall:method},
the encoding outputs, denoted as $\bm{o} = \bm{o}_1 \cdots \bm{o}_6$,
are then integrated into Seq2Seq NMT models by directly concatenated with the source input word embeddings after a linear projection.

We start with a Seq2Seq baseline with attention mechanism \cite{bahdanau2014neural} for study,
following previous studies of the same research line,
and then integrate source dependency syntax by SAWRs.
We conduct experiments on Chinese-English and English-Vietnamese translation tasks, respectively.
The results show that our method is very effective in source syntax integration.
With source dependency syntax, the performances of Chinese-English and English-Vietnamese translation
can be significantly boosted by 1.74 BLEU points and 0.80 BLEU points, respectively.
%The results show that our proposed approach is slightly more effective than a Tree-RNN baseline,
%and meanwhile our method is more efficient.
We also compare the method with the representative Tree-RNN and Tree-Linearization approaches of syntax integration,
finding that our method is able to achieve larger improvements than the two approaches for both tasks.
%We conduct several experimental analyses on the Chinese-English dataset in order to better understand our proposed approaches.
All the codes are released publicly available at https://github.com/zhangmeishan/SYN4NMT under Apache License 2.0.

%adapted from \newcite{chen-EtAl:2017:Long6} \newcite{li-EtAl:2017:Long}.
%Surprisingly, although Tree-Linearization can bring better better performances for Chinese-English translation,
%it achieves little gain for English-Vietnamese translation.

%Further, we verify the proposed approach on a sequence-to-sequence Transformer model,
%which is a stronger baseline system for NMT.
%Experimental results demonstrate that the overall tendency is similar.
%The SAWR model is able to obtain improvements of 1.04 and 0.87 BLEU points for Chinese-English and English-Vietnamese translation, respectively.
%We conduct several experimental analysis on the Chinese-English dataset in order to better understand our proposed approaches.
%All codes will be publicly available at http://github.com/xxx under Apache License 2.0.

%Our two approaches involve only input changes in order to integrate source syntax.

%All our codes will be released at \url{https://github/xyz}  under Apache License 2.0.

\section{Baseline}
We take the simple yet effective Seq2Seq model with attention mechanism proposed by \newcite{luong-pham-manning:2015:EMNLP} as our baseline.
Under the standard encoder-decoder architecture,
an encoder first maps the source-language input sentence % (e.g., $x = x_1 \cdots x_n$)
into a sequence of hidden vectors, % (e.g., $\bm{h} = \bm{h}_1 \cdots \bm{h}_n$),
and a decoder then incrementally predicts the target output sentence. % (e.g., $y = y_1 \cdots y_m$).
%Figure \ref{fig:model-framework} shows the framework of the baseline.
In particular, we should notice that several recent models \cite{vaswani2017attention,zheng2017modeling,yongcheng:2018:ACLMain}
which have been shown to be more powerful can also serve
as our baseline,
since these models focus on very different aspects of NMT,
which could be potentially complementary with our focus of syntax integration.
We will demonstrate it by experimental analysis as well.

%\begin{figure}[tb]
%	\centerline{\includegraphics[scale=1.1]{figures/figure2-seq2seq.pdf}}
%	\caption{The baseline model, where the next word $j$ is predicted. }\label{fig:model-framework}
%\end{figure}

\subsection{Encoder}
In the encoder part,
a single-layer bi-directional recurrent neural network (Bi-RNN) is employed to encode the sentence in order to
capture features from the current word and the unbounded left and right contextual words.
Given a source-language input sentence $\bm{x} = x_1 \cdots x_n$ and its embedding sequence $\bm{e}^{x_1} \cdots \bm{e}^{x_n}$,
the Bi-RNN produces an encoding sequence of dense vectors $\bm{h} = \bm{h}_1 \cdots \bm{h}_i  \cdots \bm{h}_n$:
\begin{equation}
\begin{split}
&\bm{h}_i =  \overrightarrow{\bm{h}}_i \oplus \overleftarrow{\bm{h}}_i,  \\
& \overrightarrow{\bm{h}}_i = \texttt{rnn}^{\textup L}(\bm{e}^{x_i}, \overrightarrow{\bm{h}}_{i-1}) \\
& \overleftarrow{\bm{h}}_i =  %\text{$=$}
\texttt{rnn}^{\textup R}(\bm{e}^{x_i}, \overleftarrow{\bm{h}}_{i+1})  \\
\end{split}
\end{equation}
where
$\texttt{rnn}^{\textup L/R}$ can be either GRU \cite{cho-EtAl:2014:EMNLP2014} or LSTM.
We use GRU all through this paper for efficiency following \newcite{chen-EtAl:2017:Long6}.

\subsection{Decoder}
The decoder part incrementally predicts the target word sequence $\bm{y} = y_1 \cdots y_m$, whose translation probability is defined as follows:
\begin{equation}
p(\bm{y}|\bm{x}) = \prod_{j=1}^{m} p(y_j | y_1 \cdots y_{j-1}, \bm{h}).
\end{equation}
The training objective is to maximize the probability of the reference translation.
During evaluation, we aim to search for a target sentence with the highest probability for a given source sentence.

The probability of the $j$-th target word is computed by a two-layer feed-forward neural network:
\begin{equation} \label{decode-ref}
p(y_j | y_1\cdots y_{j-1}, \bm{h}) = g(\bm{s}_{j-1},\bm{c}_j),
\end{equation}
where $\bm{s}_{j-1} = \texttt{rnn}^{\text{tgt}}(\bm{e}^{y_{j-1}}\oplus \bm{c}_{j-1}, \bm{s}_{j-2})$ is the output of a left-to-right RNN  over the predicted words, and the $\bm{c}_j$/$\bm{c}_{j-1}$ is the weighted sum over the encoding sequence $\bm{h}$ of the source sentence via the attention mechanism,
which is computed as follows:
\begin{equation} \label{attention-score}
\begin{split}
\bm{c}_j &= \sum_{k=1}^{n} \alpha_{j,k}\bm{h}_k \\
\alpha_{j,k} & = \frac{\exp(\beta_{j,k})}{\sum_{l=1}^{n} \exp(\beta_{j,l}) } \\
\beta_{j,l} &= \bm{s}_{j-1}^{\textup T} \bm{W}^a \bm{h}_l
\end{split}
\end{equation}
where $\bm{W}^a$ is the model parameter in attention.

\section{Our Method}
Syntax information has been demonstrated to be valuable for NMT.
Previously, there were two representative approaches to encode syntax into an NMT model.
The first approach directly represents an input syntax tree by \textbf{Tree-RNN},
and then uses the Tree-RNN outputs as additional encoder inputs for NMT.
The second approach models source syntax trees indirectly
by first converting a hierarchical tree into a sequence of symbols,
and then use the symbols as inputs for NMT.
The second method is referred to as \textbf{Tree-Linearization} here.

Tree-RNN is able to represent the syntax structures fully and comprehensively.
However, because of the heterogeneity of different syntax trees,
this approach suffers serious inefficiency problem as the increased difficulty
of batch computation for GPU neural computation.
%\footnote{We can acquire a batch of homogeneous tensor calculations by explicit position indexes for sequential inputs,
%while the acquisition is much more difficult for heterogenous tree-structural inputs. }
The second approach exploits an alternative sequence to substitute the original trees,
which solves the inefficiency problem.
But it may bring loss of syntax information
because the hierarchical tree structure is no longer maintained in the new representation,
which could be potentially useful for NMT.

Both the two syntax integration approaches are based on discrete 1-best outputs of a supervised dependency parser,
which may suffer from the error propagation problem.
Incorrect syntax trees as inputs for NMT may produce erroneous outputs,
leading to inappropriate translation results.
In order to alleviate the problem,
we present a novel method not using the discrete parsing outputs.

\begin{figure}[tb]
	\centerline{\includegraphics[scale=0.7]{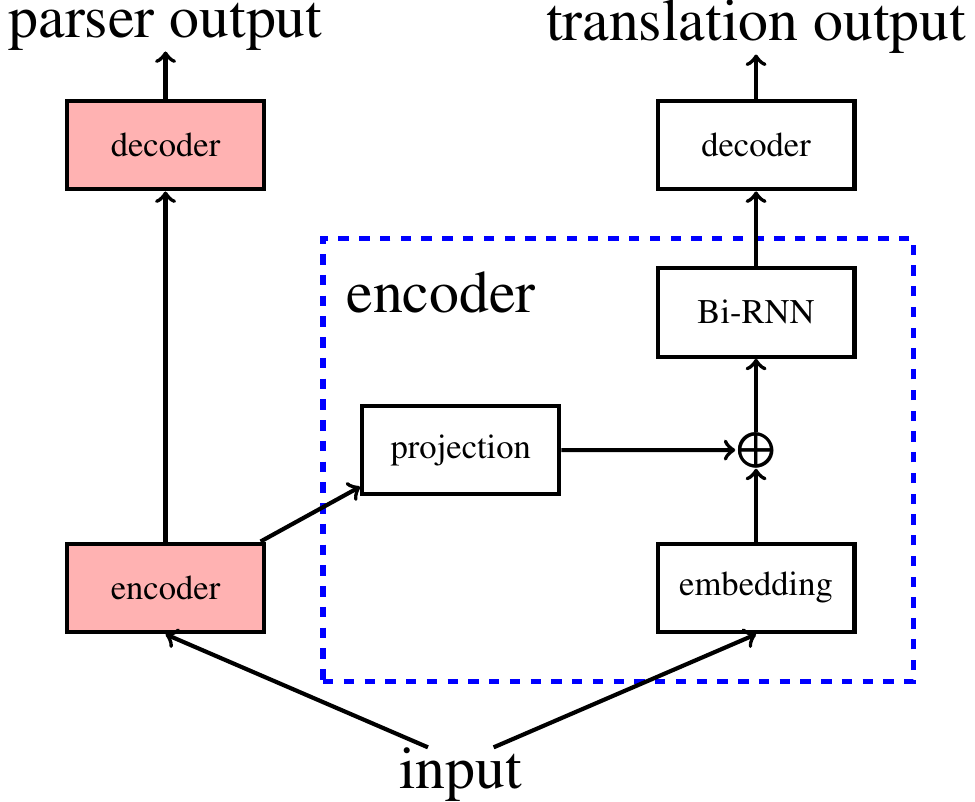}}
	\caption{The framework of the SAWR approach, where the left part shows the encoder-decoder of a supervised dependency parsing model and the right part shows the NMT encoder-decoder.  }\label{model-extractor}
%\vspace{-0.2cm}
\end{figure}

%since intuitively only certain kinds of syntactic relations are related with the translation task,
%while others could be noise.
%They adopt a multi-task learning architecture to produce latent dependency graphs which are jointly optimized with NMT.
%Their results show that improved performances can be obtained by pretrained parameters from a supervised dependency parser,
%while to our strange, without joint optimization their integration of dependency syntax gives even worse performances by using the discrete outputs directly.

We focus on supervised dependency parsing models which can be formalized as an encoder-decoder architecture,
and exploit the encoder outputs as the inputs for our Seq2Seq NMT model.
The encoder outputs are sequences of dense vectors aligning with the source sentential words,
as shown in Figure \ref{fig:overall:method},
and thus they could be easily combined with the encoder part of our NMT model.
We refer to this method as \textbf{SAWR} for brief.
Our approach takes the implicit hidden outputs from a supervised parser as inputs for NMT,
which greatly reduces the direct influence brought from discrete 1-best parser outputs.

%Actually, the neural structure of a deep encoder-decoder dependency parser
%have several layers from the bottom input words to the top output trees.
%Figure the neural structure of the BiAffine dependency parser proposed by,
%which has achieved the top performance in the literature.
%As the neural layer goes closer to the output,
%the implicit hidden features are more directly related with syntax.
%The exploration of the final output is just one extreme case.

Figure \ref{model-extractor} shows the framework of SAWR.
Concretely, we first project the encoder outputs of a dependency parsing model
into a sequence of vectors by a feed-forward linear layer,
as shown by the projection module in Figure \ref{model-extractor}:
\begin{equation}\label{hidden-project}
 \bm{s}_i =  \bm{W}\bm{o}_i  + \bm{b}
\end{equation}
where $\bm{o} \text{$=$} \bm{o}_1 \cdots\bm{o}_n$ is the encoder output of a parsing model,
$\bm{W}$ and $\bm{b}$ are model parameters.
Then we concatenate the resulting vectors with the source embeddings as inputs for the baseline Bi-RNN Encoder.
Thus the encoder process can be formalized as follows:
\begin{equation}\label{hidden-combine}
\bm{h} = \text{Bi-RNN} \big( \bm{e}^{x_1}\oplus\bm{s}_1,
\cdots,  \bm{e}^{x_n}\oplus\bm{s}_n\big).
\end{equation}

Noticeably, the SAWR method can be regarded as an adaption of joint learning as well.
We can train both dependency parsing and machine translation model parameters concurrently.
In this work, we focus on the machine translation task
and do not involve the training objective of dependency parsing.
However, we can still fine-tune model parameters of the encoder part of dependency parsing
by back-propagating the training losses of NMT into this part as well.

Actually, SAWRs are also similar to the ELMO embeddings \cite{N18-1202-elmo}.
ELMO learns context word representations by using language model as objective,
while SAWRs learn syntax-aware word representations by using dependency parsing as objective.
On the other hand,
compared with the Tree-RNN and Tree-Linearization methods which encode syntax trees by neural networks directly,
SAWRs are less sensitive to the output syntax trees.
Thus the SAWR method can alleviate the error propagation problem.

%The additional cost of this integration is the lowest among the three methods,
%because the decoder part of dependency parsing is saved, and the added part is less the save one.

\section{Experiments}

\subsection{Settings}

\textbf{Data.} We conduct experiments on the Chinese-English and English-Vietnamese translation tasks, respectively.
For Chinese-English,
we use the parallel training data from the publicly available LDC
corpora,\footnote{LDC2002E18, LDC2003E07, LDC2003E14, Hansards portion of LDC2004T07, LDC2004T08 and LDC2005T06.}
with 28.3M Chinese words and 34.5M English words, respectively,
consisting of 1.25M sentence pairs,
and test model performances on the NIST datasets,
using NIST MT02 as the development data, and MT03-06 as test datasets.
For English-Vietnamese, we use the standard  IWSLT 2015 dataset,\footnote{https://nlp.stanford.edu/projects/nmt/}
which consists of about 133K sentence pairs,
and evaluate our models by exploiting the TED tst2012  and tst2013
as the development and test datasets, respectively.

For the source side sentences,
we construct vocabularies of the most frequent 50K words,
while for the target side sentences,
we apply byte-pair encodings (BPE) \cite{sennrich-haddow-birch:2016:P16-12}
with 32K merges to obtain subword units,
and construct the target vocabularies by the most frequent 32K subwords.
During training,
we use only the sentence pairs whose source and target lengths both are no longer than $50$ and $150$ for Chinese-English and English-Vietnamese translations, respectively.

\textbf{Evaluation.}
We use the case insensitive 4-gram BLEU score as the main evaluation metrics  \cite{papineni-EtAl:2002:ACL},
and adopt the script \texttt{multi-bleu.perl} in the Mose toolkit.\footnote{http://www.statmt.org/moses}
Significance tests are conducted based on the best-BLEU results for each approach by using bootstrap resampling \cite{koehn:2004:EMNLP}.

Alternatively, in order to compare the effectiveness of our model with other syntax integration methods,
we implement a \textbf{Tree-RNN} approach and a \textbf{Tree-Linearization} approach,
respectively:
\begin{itemize}
  \item Tree-RNN: We build a one-layer bi-directional Tree-RNN with GRU over input word embeddings,
  producing syntax-enhanced word representations, which are then fed into the encoder of NMT as basic inputs.
  The method is similar to the model proposed by \newcite{chen-EtAl:2017:Long6}.
  \item Tree-Linearization: We first convert dependency trees into constituent trees \cite{weiweisuntacl13}, and then feed it into the NMT model proposed by \newcite{li-EtAl:2017:Long}.
\end{itemize}

\textbf{Hyperparameters.}
We set the dimension sizes of all hidden neural layers to 1024,
except the input layers for RNNs (i.e. input word embeddings and the projection layer of SAWR),
which are set to 512.
We initialize all model parameters by random uniform distribution between $[-0.1, 0.1]$.
We apply dropout on the output layer of word translation with a ratio of 0.5.

We adopt the Adam algorithm \cite{kingma2014adam} for parameter optimization, with the initial learning rate of $5\times 10^{-4}$,
the gradient clipping threshold of 5, and the mini-batch size of 80.
During translation, we employ beam search for decoding with the beam size of 5.
%In each epoch, we evaluate the model on the development data, after consuming  all the training data, which amounts to about 14K model updates.
%We decay the learning rate by $0.75$ when the development performance decreases compared with the last epoch.

%\renewcommand\arraystretch{0.90}
\setlength{\tabcolsep}{6pt}
\begin{table*}[!ht]
\begin{center}
\begin{tabular}{c|cccc|c}
\hline
System & MT03  & MT04  & MT05  & MT06 & Average/$\Delta$ \\ \hline
%\multicolumn{6}{c}{ \bf Single Model }    \\ \hline \hline
Baseline &    36.44     &   39.35    &   36.26    &  36.32     &   37.09  \\ \hline
\textbf{SAWR} &       \bf 38.42   &   \bf 40.60    &   \bf 38.27    &  \bf 38.04    &  \bf 38.83/+1.74 \\ \hline\hline
Tree-RNN  &   38.12   &     40.35  &    37.86   &   37.32   &  38.41/+1.32    \\
Tree-Linearization &    37.95    &   40.24    &  37.64   &  37.44    &  38.32/+1.23
    \\ \hline \hline
\multicolumn{6}{c}{ Previous Work}    \\ \hline \hline
\newcite{chen-EtAl:2017:Long6}  &    35.64 & 36.63 & 34.35 & 30.57  & 34.30/\textbf{+2.59}    \\
\newcite{li-EtAl:2017:Long}  &    34.9 &  38.6 &   \bf 35.5 & \bf 35.6 &  \textbf{36.15}/+1.45    \\
\newcite{chen-EtAl:2017:EMNLP20173}  &    \bf 35.91 &   \bf 38.73 &  34.18 &   33.76 & 35.65/+1.52   \\
\hline
\end{tabular}
\caption{Final results of Chinese-English translation.
All syntax-integrated approaches are significantly better than the baseline system ($p < 0.05$). } \label{table:chinese:result}
\end{center}
%\vspace{-0.3cm}
\end{table*}

\textbf{Source-Side Parsing.}
We employ the state-of-the-art BiAffine dependency parser recently proposed by \newcite{dozat2016deep} to obtain the source-side dependency syntax information.
The BiAffine parser can also be understood as an encoder-decoder model,
where the encoder part is a three-layer bi-directional LSTM over the input words,
and the decoder uses BiAffine operations to score all candidate dependency arcs and finds the highest-scoring trees via dynamic programming.

For Chinese-English translation, we train the dependency parser on Chinese Treebank 7.0 with Stanford dependencies,\footnote{https://nlp.stanford.edu/software/stanford-dependencies.shtml}
using 50K random sentences as the training data and the remaining as the test data.
The parser achieves $81.02\%$ parsing accuracy (labeled attached score, LAS) on the test dataset.
For English-Vietnamese translation, we train the dependency parser on English WSJ corpus,
following the same data split as \newcite{dozat2016deep},
and obtaining a LAS of $93.84\%$ on the test dataset.\footnote{For simplicity,
we use only words as inputs for both Chinese and English dependency parsing, avoiding the influences brought by other inputs, such as automatic POS tags.}

\subsection{Speed Comparison}
All our experiments are run on a single GPU NVIDIA TITAN Xp.
We report the averaged one-epoch training time on the Chinese-English translation dataset (consuming all 125M sentence pairs)
as follows:

%Here we only report the training time for fair comparison.
%\setlength{\tabcolsep}{0pt}
\begin{center}
%\begin{small}
%	\vspace{+.3em}
\begin{tabular}{ c  c   }
\hline
 Baseline & 105 min \\ \hline
 \textbf{SAWR} & 142 min \\ \hline \hline
 Tree-RNN & 498 min \\
 Tree-Linearization  &  137 min \\
\hline
\end{tabular}
%	\vspace{+.3em}
%\end{small}
\end{center}
The SAWR system spends averaged 142 minutes,\footnote{We exclude the time consumed by the encoder part of the dependency parsing model for fair comparisons, as other methods require to perform parsing in an offline way.}
37 minutes slower than the baseline model.
The Tree-Linearization spends averaged 137 minutes per epoch,
which is the fastest syntax integration method.
Our SAWR approach spends 5 more minutes than Tree-Linearization,
appropriate 3.5\% of the total spend time per epoch,
which could be negligible.
%Thus the two methods are comparable in efficiency.
The Tree-RNN model spends 498 minutes per epoch,
nearly four times slower than the baseline model.\footnote{The Tree-RNN model is implemented with deliberate batching motivated by \newcite{neubig17nips}, without which the model is intolerably slow, reaching about 1,900 minutes per epoch.}
According to  the results,
we can conclude that the Tree-RNN model is highly inefficient for encoding dependency syntax,
whereas the SAWR and Tree-Linearization are almost as efficient as the baseline Seq2Seq system.

\subsection{Main Results}
\subsubsection{Chinese-English Translation}
Table \ref{table:chinese:result} shows the main results of all approaches on Chinese-English datasets.
Considering the effect of random initialization, we train three individual models for each approach,
and use the averaged BLEU scores for fair comparisons.

According to the results, we can see that all syntax-integrated approaches can bring significant improvements over the baseline system,
which denotes that syntax is highly effective for Chinese-English machine translation.
In addition, the proposed SAWR approach obtains the largest BLEU improvements,
averaged $\Delta =1.74$ BLEU points better than the baseline system.
The Tree-RNN and Tree-Linearization approaches bring improvements of averaged $\Delta =1.32$  and
$\Delta =1.23$ BLEU points, respectively.
The results show that our implicit syntax-aware encoding method is better than Tree-RNN and Tree-Linearization.

%Besides the single-model approach, we also report the results of three-model  ensemble approaches for in-depth study \cite{denkowski17nmt}.
%Further, we conduct ensemble experiments to compare our proposed approaches under stronger baselines \cite{denkowski17nmt}.
%On the one hand, we can make our results more trustable.
%On the other hand, we aim to see the difference of the three proposed systems.
%We follow \newcite{denkowski17nmt} to perform model ensemble by averaging the output scores of  individual models.
%As shown in Table \ref{table:chinese:result},
%model ensemble is able to improve the translation quality effectively,
%which is consistent with the previous findings in \cite{zhou-EtAl:2017:Short1,denkowski17nmt}.
%After homo-approach three-model ensemble, syntax information can still bring better performances,
%resulting in averaged BLEU improvements of $42.09-41.24=0.84$ by Tree-RNN,
%$42.45-41.24=1.21$ by Tree-Linearization and $42.60-41.24=1.36$ by SAWRs, respectively.

We compare our NMT models with other state-of-the-art methods as well.
The results are just for reference since experimental details could be very different.
In particular, we list the relative improvements over the corresponding baseline models by integrating syntax structures, which are calculated according to their papers.
All these studies exploit lower baselines compared with our models.
The Tree-RNN and Tree-Linearization are essentially similar to \newcite{chen-EtAl:2017:Long6} and  \newcite{li-EtAl:2017:Long}, respectively.
As shown, our approaches can still obtain large improvements based on a stronger baseline.

\setlength{\tabcolsep}{6.0pt}
\begin{table}[t]
\begin{center}
\begin{tabular}{c|c}
	\hline
	System & tst 2013 / $\Delta$  \\ \hline
	Baseline &   28.29  \\ \hline
    \textbf{SAWR} &    \bf 29.09/+0.80       \\
	Tree-RNN &   28.51/+0.22   \\
    Tree-Linearization &  28.93/+0.64      \\
    \hline
\end{tabular}
\caption{Final Results on the IWSLT 2015 English-Vietnamese translation task. Only SAWR is significantly better than the baseline system ($p < 0.05$). }
\label{table:germany:result}
\end{center}
%\vspace{-0.3cm}
\end{table}

\subsubsection{English-Vietnamese Translation}
Table \ref{table:germany:result} shows the final results on the IWSLT 2015 English-Vietnamese translation task.
The overall tendency is similar to that of Chinese-English translation.
The syntax information can boost the translation performances by using any of the three approaches.
The SAWR approach gives the best translation performance,
significantly outperform the baseline system by $\Delta = 0.80$ BLEU points.
While although the other two approaches bring better performances,
the improvements are not significant.
The results demonstrate the advantage of the proposed implicit SAWR approach.
By not using the 1-best parser outputs,
our approach can reduce the error propagation problem, thus bring larger improvements with syntax.

In particular, we find that the increases of BLEU scores are smaller than that of Chinese-English translation
by integrating syntactic features.
The averaged BLEU increases are $0.55$ for English-Vietnamese
and $1.43$ for Chinese-English.
The possible reason may be due to that the source English sentences are more grammatically rigorous than Chinese sentences.
For example, the English functional words such as ``of'' and ```s''
which indicate the possessive relationship,  should be always kept in sentences by standard,
while their Chinese correspondence ``的'' may be omitted in sentences.
%Thus the syntax structural information provides less extra information for source English sentences,
%which makes syntax less useful for NMT.

\subsection{Analysis}

In this section, we conduct analysis on Chinese-English translation from different aspects
to better understand the SAWR approach of integrating source-side dependency syntax for NMT.

\setlength{\tabcolsep}{3.0pt}
\begin{table}[t]
	\begin{center}
		%\begin{small}
			\begin{tabular}{c|cccc|c}
				\hline
				Parser &  MT03  & MT04  & MT05  & MT06 & Average \\ \hline
				no Tune & \bf 38.42  &   \bf 40.60  &  \bf 38.27  &  \bf 38.04   & \bf 38.83 \\
				Tune & 37.33 & 39.45 & 36.93 & 37.03 & 37.69   \\ \hline
                %None & \multirow{2}{*}{36.44} & \multirow{2}{*}{39.35} & \multirow{2}{*}{36.26} & \multirow{2}{*}{36.32} & \multirow{2}{*}{37.09}   \\
                %(Baseline) &   &   &   &   &     \\ \hline
			\end{tabular}
			\caption{The influence of fine-tuning parser parameters in the SAWR system. }
			\label{table:analyze:tune}
		%\end{small}
	\end{center}
%\vspace{-0.3cm}
\end{table}

\setlength{\tabcolsep}{6pt}
\begin{table*}[ht]
\begin{center}
\begin{tabular}{c|cccc|c}
\hline
System & MT03  & MT04  & MT05  & MT06 & Average/$\Delta$ \\ \hline
Baseline$\times$3  &     40.90   &  43.25     &  40.64     &   40.16  & 41.24      \\ \hline
\textbf{SAWR}$\bm{\times}$\textbf{3} &     41.94    &   44.59    &   41.91    &    41.97  &    42.60/+1.36   \\
Tree-RNN$\times$3 &    42.03    &   44.15    &   41.50    &   41.41   &  42.27/+1.03      \\
Tree-Linearization$\times$3 &     41.74    &   44.23    &   41.32    &   41.44   &  42.18/+0.94    \\ \hline
\textbf{Hybrid} &    \bf 42.72   &    \bf 45.14   &    \bf 42.38   &   \bf 42.15   &  \bf43.10/+1.86     \\ \hline
\end{tabular}
\caption{Ensemble performances, where the Hybrid model denotes SAWR + Tree-RNN + Tree-Linearization. } \label{table:chinese:ensemble:result}
\end{center}
%\vspace{-0.1cm}
\end{table*}

\subsubsection{Fine-Tuning Syntax-Oriented Inputs}
The SAWR approach directly uses the encoder outputs of a dependency parser as extra inputs for NMT.
In the above experiments, we keep the parser model parameters fixed, letting them uninfluenced from NMT optimization.
Actually, this part can be further fine tuned along with the NMT learning,
by treating them as one kind model parameters.
Thus there arises a question that whether fine-tuning the parser model parameters can bring better performance.

As an interesting attempt, we can simultaneously
fine tune the parameters of both the parser and the Seq2Seq NMT model during training.
Figure \ref{table:analyze:tune} shows the results.
We can see that fine-tuning decreases the average BLEU score by $38.83-37.69=1.14$ significantly.
This may be because that fine-tuning disorders the representation ability of
the parser and makes its function more overlapping with other network components.
This further demonstrates that pretrained syntax-aware word representations are helpful for NMT.

\subsubsection{\normalsize Alignment Study}
Alignment quality is an important metric to illustrate and evaluate machine translation outputs.
Here we study how syntax features influence the alignment results for NMT.
We approximate the alignment scores by the attention probabilities
as shown in Equation \ref{attention-score}.\footnote{We aim to offer an intuitive interpretation by a carefully-selected example. In fact, the alignment computation method here may be problematic \cite{W17-3204}.}
For better understanding the effectiveness of syntax,
we choose the target-side English word ``of'' for comparison,
which is a grammatical functional word.

\begin{figure}[tb]
	\centerline{\includegraphics[scale=0.79]{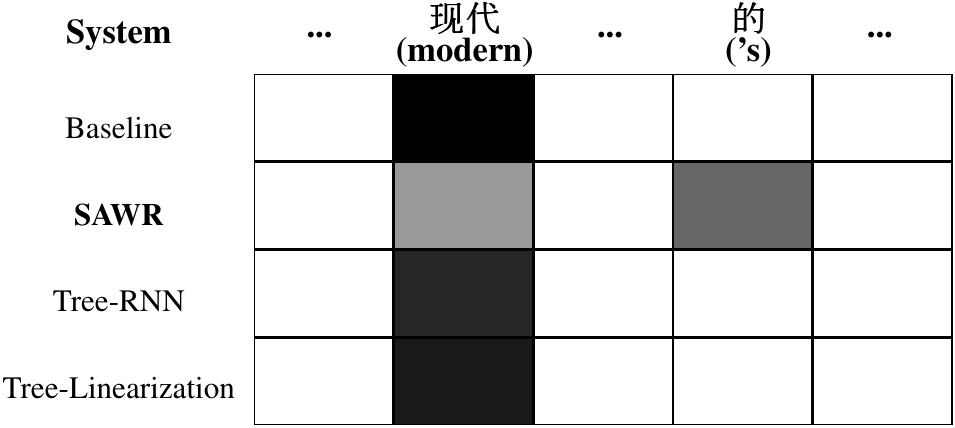}}
	\caption{Alignments for the baseline and syntax-integrated systems,
		where the same example in Figure \ref{fig:overall:method} is analyzed and the target English word is ``of''. }\label{analyze-linear}
%\vspace{-0.3cm}
\end{figure}

Figure \ref{analyze-linear} shows the alignment probability distributions returned by different approaches.
Intuitively, this word should be aligned with the Chinese word ``的(de)''.
But according to the results, we can see that
only the SAWR model distributes a high attention score to it, which is consistent with our intuition.
The other three models are all aligned to the source word ``现代 (modern)'' with high confidence over 85\%.
The possible reason for ``of'' being aligned to ``现代 (modern)'' could be due to that ``of modern'' is a high-frequency collocation in the training corpora.

%We examine the alignment results among the baseline system
%and the syntax-integrated SAWRs, Tree-RNN and Tree-Linearization models.

%While these words are grammatically indispensable for English,
%it is difficult in most cases to find their correspondences  in the Chinese counterpart.
%The mismatching problem could bring inaccuracies in alignments to the baseline model.
%In contrast, the syntax-integrated models should be able to find suitable alignments when translate to these words.

\subsubsection{\normalsize Ensemble Study}
Here we perform model ensembles to examine the divergences of the three syntax-integration approaches \cite{zhou-EtAl:2017:Short1,denkowski17nmt}.
Intuitively, the hetero-approach ensemble which combines three NMT models of different methods
should obtain better performances than homo-approach ensembles which combine three NMT models of the same method,
since NMT models of different syntax-integrations approaches have larger divergences.

Table \ref{table:chinese:ensemble:result} shows the results.
First, we can see that ensemble is one effective technique to improve the translation performances.
More importantly, the results show that the heterogeneous ensemble achieves averaged BLEU improvements by $43.10-41.24=1.86$ points,
better than the gains achieved by all three homo-approach ensembles,
denoting that the three approaches could be mutually complementary in representing dependency syntax,
and the resulting models of the three approaches are highly diverse.

\begin{figure}[tb]
	\centerline{\includegraphics[scale=1.0]{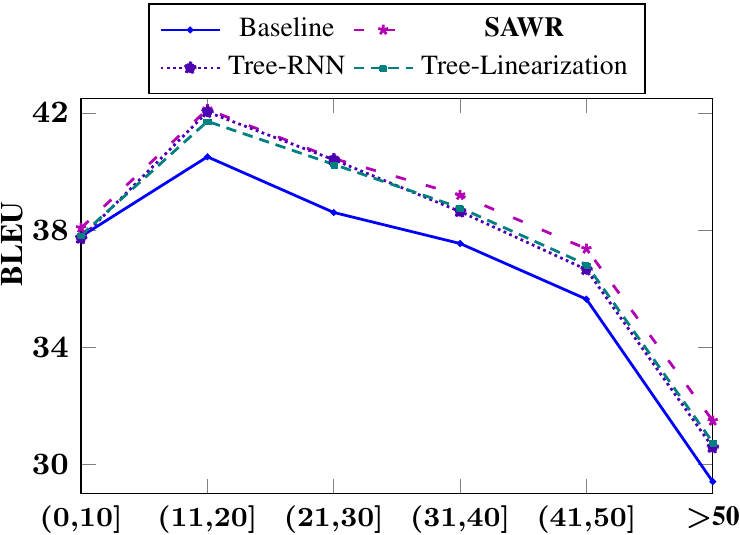}}
	\caption{The effect of source input length. }\label{analyze-source-length}
%\vspace{-0.3cm}
\end{figure}

\subsubsection{Analysis by Source Sentence Length}
Intuitively, by introducing the source syntax into the NMT model,
relations between long-distance words are explicitly modeled by dependency trees,
thus we can expect that models enhanced by source syntax are able to bring better translations for longer sentences.
Figure \ref{analyze-source-length} shows the performances of the baseline and all syntax-enriched models in terms of source sentence lengths,
where we bin all the MT03-MT06 sentences by their lengths into six intervals.
The results show that the BLEU scores are improved significantly when source sentential lengths are over 10,
which confirms our intuition.

\begin{figure}[tb]
	\centerline{\includegraphics[scale=0.90]{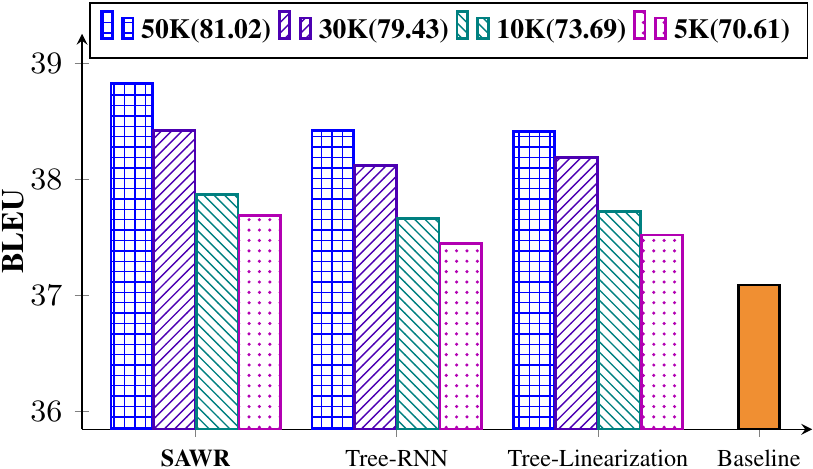}}
	\caption{The effect of dependency parsing performances on our proposed approaches. }\label{analyze-parsing-performance}
%\vspace{-0.05cm}
\end{figure}

\setlength{\tabcolsep}{6pt}
\begin{table*}[ht]
	\begin{center}
		%\begin{small}
			\begin{tabular}{c|cccc|c}
				\hline
                System & MT03  & MT04  & MT05  & MT06 & Average/$\Delta$ \\ \hline
				Transformer & 40.45  &   42.76  &  40.09  &  39.67   & 40.74  \\ \hline
                \bf SAWR & \bf 41.63 & \bf 43.60 & \bf 41.68 & \bf 40.21 & \bf 41.78/+1.04   \\ \hline \hline
				Tree-RNN & 41.24 & 43.38 & 41.04 & 40.02 & 41.42/+0.68    \\
                Tree-Linearization & 41.12 & 43.02 & 41.04 & 39.86 & 41.26/+0.52  \\ \hline
			\end{tabular}
			\caption{Final results based on the transformer. Only the SAWR results are significantly better ($p < 0.05$).}
			\label{table:transformer:result}
		%\end{small}
	\end{center}
%\vspace{-0.2cm}
\end{table*}

\subsubsection{Effect of Parsing Performance}
Finally, we examine how the performance of the dependency parser influences the final translation quality.
While the full dependency parser is trained on 50K sentences,
we retrain three weaker dependency parsers on 30K, 10K and 5K sentences, respectively.
%Overall, the BLEU score decreases accompanying with the dropping of parsing performances,
%which also verifies the effectiveness of dependency syntax.
Figure \ref{analyze-parsing-performance} shows the NMT BLEU scores and the parsing accuracies.
It is clear that the parsing accuracy directly influences the translation quality, indicating the effectiveness and importance of exploiting syntactic  information.

%Finally, we study the influence of source sentence length in translation,
%following \newcite{bahdanau2014neural}.

\subsubsection{Transformer as Baseline}
Here we conduct experiments based on the transformer NMT model \cite{vaswani2017attention},
which is a stronger baseline, to further verify the effectiveness of our proposed method.
This also demonstrates that the proposed SAWR method does not limit to a certain NMT baseline.
Concretely, we extend the bottom word representations
by incorporating syntactic encodings $\bm{s} \text{$=$} \bm{s}_1 \cdots\bm{s}_n$ (shown
in Equation \ref{hidden-project}) into them,
and then feed them into the transformer encoder by a linear projection layer to align with the input dimension.
We implement Tree-RNN and Tree-Linearization for Transformer in a similar way,
only adapting the source input word representing.
We adopt a widely-used setting with 8 heads, 6 layers and the hidden dimension size of 512.

Table \ref{table:transformer:result} shows the results.
As shown, the transformer results are indeed much better than RNN-based baseline.
The BLEU scores show an average increase of $40.74 - 37.09 = 3.65$.
In addition, we can see that syntax information can still give positive influences based on the transformer.
The SAWR approach can also outperform the baseline system significantly.
Particularly, we find that our SAWR approach is much more effective than the Tree-RNN and Tree-Linearization approaches.
The results further demonstrate the effectiveness of SAWRs in syntax integration for NMT.

\section{Related Work}
By explicitly expressing the structural connections between words and phrases,
syntax trees  been demonstrated helpful in SMT \cite{liu-liu-lin:2006:COLACL,cowan-kuucerova-collins:2006:EMNLP,marton-resnik:2008:ACLMain,xie-mi-liu:2011:EMNLP,li-resnik-daumeiii:2013:NAACL-HLT,williams2016syntax}.
Although the representative Seq2Seq NMT models are able to capture latent long-distance relations
by using neural network structures such GRU and LSTM \cite{sutskever2014sequence,wu2016google},
recent studies show that explicitly integrating syntax trees into NMT models can bring further gains \cite{sennrich-haddow:2016:WMT,shi-padhi-knight:2016:EMNLP2016,zhou-EtAl:2017:Short2,wu-EtAl:2017:Long2,aharoni-goldberg:2017:Short}.
Under the NMT setting,
the exploration of syntax trees could be more flexible,
because of the strong capabilities of neural network in representing arbitrary structures.

Recursive neural networks based on LSTM or GRU have been one natural method to model syntax trees \cite{zhu2015long,tai-socher-manning:2015:ACL-IJCNLP,li-EtAl:2015:EMNLP5,zhang-lu-lapata:2016:N16-1,teng2016bidirectional,miwa-bansal:2016:P16-1,kokkinos-potamianos:2017:EACLshort},
which are capable of representing the entire trees globally.
\newcite{eriguchi-hashimoto-tsuruoka:2016:P16-1} present the first work to apply a bottom-up Tree-LSTM for NMT.
The major drawback is that its bottom-up composing strategy is insufficient for bottom nodes.
Thus bi-directional extensions have been suggested \cite{chen-EtAl:2017:Long6,yang-EtAl:2017:EMNLP20172}.
Since Tree-RNN suffers serious inefficiency problem,
\newcite{li-EtAl:2017:Long} suggest a Tree-Linearization alternative,
which converts constituent trees into a sequence of symbols mixed with words and syntactic tags.
The method is as effective as Tree-RNN approaches yet more effective.
Noticeably, all these studies focus on constituent trees.

There have been several studies for NMT using dependency syntax.
\newcite{hashimoto-tsuruoka:2017:EMNLP2017} propose to combine the head information with sequential words together as source encoder inputs,
where their input trees are latent dependency graphs.
Recently, there are several studies by using convolutional neural structures to represent source dependency trees,
where tree nodes are modeled individually \cite{chen-EtAl:2017:EMNLP20173,bastings-EtAl:2017:EMNLP2017}.
\newcite{wu2017improved} build a syntax enhanced encoder by multiple Bi-RNNs over several different word sequences based on different traversing orders over dependency trees, i.e., the original sequential order and several tree-based orders.
All these methods require certain extra efforts to encode the source dependency syntax over a baseline Seq2Seq NMT.

%\newcite{zhang-zhang-fu:2017:EMNLP2017} firstly propose to use hidden vector features pretrained from a dependency parser
%as a kind of syntax representation method.
%They apply it on an end-to-end relation extraction model,
%showing that it can achieve comparable improvements with Tree-LSTM.
%Similarly, \newcite{gao-zhang-xiao:2017:I17-1} apply the same method to targeted sentiment classification.
%In this work, we apply the method to NMT,
%and further compare it with other syntax representation methods.

\section{Conclusion}
We proposed a novel syntax integration method, SAWR, to incorporate source dependency-based syntax for NMT.
It encodes dependency syntax implicitly, not requiring discrete syntax trees as inputs.
Experiments showed that the method can bring significantly better performances for both Chinese-English and English-Vietnamese translation tasks.
In addition,
we compared the method with two approaches based on Tree-RNN and Tree-Linearization, which has been previously exploited for syntax integration,
finding that our method is more effective and meanwhile very efficient.
%We conducted model ensemble experiments, on the one hand to demonstrate the effectiveness of our methods with stronger baselines,
%and on the other hand to show the diversity of the different syntax integration methods.
We conducted several experimental analyses to study our proposed methods deeper.

\section*{Acknowledgments}
We thank all anonymous reviewers for their valuable comments.
We thank Huadong Chen, Haoran Wei and Zaixiang Zheng for their help in implementing
baseline neural machine translation models.
This work is supported by National Natural Science Foundation of China (NSFC) grants 61525205, U1836222, and 61672211.

%\newpage
\bibliography{reference}
\bibliographystyle{acl_natbib}
\end{CJK}

\end{document}